\pdfoutput=1

\documentclass[11pt]{article}

\usepackage[ruled,linesnumbered]{algorithm2e}

\usepackage{emnlp2021}

\usepackage{times}
\usepackage{latexsym}

\usepackage[T1]{fontenc}

\usepackage{graphicx}
\usepackage{subfig}
\usepackage{float}

\usepackage[utf8]{inputenc}

\usepackage{microtype}

\usepackage{natbib}
\setcitestyle{numbers,square}

\title{Discovering Patterns of Definitions and Methods from Scientific Documents}

\begin{document}

\author{Yutian Sun \and Hai Zhuge \\
        {The Key Laboratory of Intelligent Information Processing, ICT, CAS, Beijing, China} \\ {The Great Bay University, Dongguan, China}}

\maketitle
\begin{abstract}
The difficulties of automatic extraction of definitions and methods from scientific documents lie in two aspects: (1) the complexity and diversity of natural language texts, which requests an analysis method to support the discovery of pattern; and, (2) a complete definition or method represented by a scientific paper is usually distributed within text, therefore  an effective approach should not only extract single sentence definitions and methods but also integrate the sentences to obtain a complete definition or method. This paper proposes an analysis method for discovering patterns of definition and method and uses the method to discover patterns of definition and method.  Completeness of the patterns at the semantic level is guaranteed by a complete set of semantic relations that identify definitions and methods respectively. The completeness of the patterns at the syntactic and lexical levels is guaranteed by syntactic and lexical constraints.  Experiments on the self-built dataset and two public definition datasets show that the discovered patterns are effective. The patterns can be used to extract definitions and methods from scientific documents and can be tailored or extended to suit other applications.
\end{abstract}

\section{Introduction}

Some syntax and lexicon play an important role in representing definitions and methods but others do not. To reduce the complexity of analyzing texts, the input sentences are preprocessed by the following steps that do not change the important relations (such as part-whole relation and synonymy) that render definitions and methods:

(1)	The compound sentence is split into two or more main clauses by finding the coordinators ( <coordinator> ::= "and" | "or" | "not only" | "but also" | "both" | "either" | "yet" | "but" | “as well as”) and then judging whether the content appearing before and after the coordinator matches the pattern of main clause, e.g., “\textit{self-attention is used in a variety of tasks \underline{and} is an attention mechanism.}” is split into “\textit{self-attention is used in a variety of tasks}” and “\textit{self-attention is an attention mechanism}” because this does not change the representation of definition and method.

(2)	The active voice of a predicate verb followed by a noun phrase and a prepositional phrase is transformed into its passive voice to avoid the situation that identifier is separated, e.g., “\textit{… \underline{add} a regularization loss to the original objective}” with “\textit{add}” as predicate verb is transformed into “\textit{a regularization loss \underline{is added to} the original objective}”. This is to unify different forms of representation into one form because most definitions and some methods are defined by passive voice.

(3)	All modals, adverbs (except the ones expressing negation, e.g., “not”) appearing before or after verbs and adjectives, and auxiliary “have” are removed, e.g., “\textit{philosophy \underline{can} \underline{also} be defined as an attempted science of common sense}” is transformed into “\textit{philosophy be defined as an attempted science of common sense}”. This ignores the words expressing the degrees of certainty (“definitely”) and necessity (“must”), the tense or aspect of verbs, e.g., modal “will” and auxiliary “have”.

(4)	All the nouns, adjectives and verbs (except the participle verbs modifying nouns, e.g., “\textit{proposed}” in “\textit{proposed method}”) are lemmatized and lowercased (e.g., “is” is replaced by “be”). This ignores the singular and plural of nouns and verbs, the comparative degree of adjectives because this does not influence the identification of definition and method.

(5)	Synonyms of all nouns, verbs and adjectives are searched from the WordNet \citep{0Introduction} (e.g., the synonyms of “kind” are “class”, “type”, and “sort”) to generalize the sentence without changing its original meaning.

\section{Extracting Definition Pattern Within Sentence}
\subsection{Basic Method}
The basic method consists of the following steps:

1.	Use a small set of examples of definition within selected documents to \textit{generalize initial patterns} of the definition according to basic syntax patterns, e.g., <subject> <verb> <object> is generalized from “\textit{Bayesian network is a generative model.}”. 

2.	Find the part-of-speech of the subject and object to \textit{specialize the patterns}, e.g., <adjective> <noun> “is a” <adjective> <noun> is a specialization of <subject> <verb> <object>.

3.	Generalize pattern by \textit{using phrase and clause to replace the part-of-speech}, e.g., <adjective> <noun> “is a” <adjective> <noun> can be generalized as <noun phrase> “is a” <noun phrase>.

4.	Use more examples to \textit{find a different abstraction} on the representation of object, e.g., <noun phrase> “is that” <main clause> is generalized from “\textit{The basic idea of OBD is that it is possible to take a perfectly reasonable network, delete half of the weights and wind up with a network that works just as well, or better.}”, and the <main clause> is an abstraction different from <noun phrase>, therefore forming a more general pattern: (<noun phrase> “is a” <noun phrase>) | (<noun phrase> “is that” <main clause>).

5.	\textit{Generalize the verb} by lowercasing and lemmatizing them as identifier, e.g., generalize “is” as “be” to form a pattern: <noun phrase> “be a” <noun phrase>, where “be a” is the identifier.  This is implemented in preprocessing texts.

6.	\textit{Find equivalent patterns} so that the patterns can cover more examples, e.g., the pattern <noun phrase> “be made up of” <noun phrase> and the pattern: <noun phrase> “make up” <noun phrase>.

7.	\textit{Merge the patterns} that include only different representations of identifier so that the patterns can be generalized. For example, <noun phrase> “be make up of” <noun phrase> and <noun phrase> “consist of” <noun phrase> can be merged as: <noun phrase> (“be make up of” | “consist of”) <noun phrase> to cover more cases.

8.	\textit{Test precision and recall of the patterns by using multiple datasets}, and make recommendation by making comparison with the baselines.

\subsection{The General Pattern}
\subsubsection{High-Level Pattern}
A definition is identified by one or more word(s) (called identifier, e.g., “be a”, “denote”, “be view as”, “include”, “be compose of”, “be split into”) that connects a noun phrase (i.e., the definiendum, in short DFD) to a representation (i.e., the definiens, in short DFN) that defines the meaning of the noun phrase.

The high-level pattern of definition is as follows:

<\textbf{definition}> ::= (<noun phrase> (“be” | “have”) <noun phrase>) | (<noun phrase> <other identifier> (<noun phrase>| <nominal clause> | <verb phrase> | <main clause>)) | ((<noun phrase> | <nominal clause> | <participle phrase>) <other identifier> <noun phrase>) | (<other identifier> <noun phrase> (<relative clause> | <noun phrase>))

Identifier is a word or phrase that identifies the definition within a sentence. It consists of auxiliary “be”, lexical verb “have” and other identifiers consisting of subclass identifier, synonymy identifier and part-whole identifier. The “be” and “have” are the most frequently used identifiers.  The pattern of the other identifier is as follows:

<other identifier> ::= <subclass identifier> | <part-whole identifier> | <synonymy identifier>

The complete patterns are shown in Appendix A.

The following are five examples of the pattern.

Example 1: “[\textit{Self-attention, sometimes called intra-attention}], \underline{\textit{is an}} \{\textit{attention mechanism relating different positions of a single sequence in order to compute a representation of the sequence}\}.”, where the content within "[ ]" marks the definiendum, and the content within "\{ \}" marks the definiens, the content underlined "\_" marks the identifier. It is a special case of a pattern: <noun phrase> “be an” <noun phrase>.

Example 2: “[\textit{YouTube-VOS, with diverse scenes}], \underline{\textit{consists of}} \{\textit{3471, 474, and 508 video clips for training, validation, and test, respectively}\}”, which is a special case of the pattern: <noun phrase> “consist of” <noun phrase>.

Example 3: “\textit{From a probabilistic perspective}, [\textit{translation}] \underline{\textit{is equivalent to}} \{\textit{finding a target sentence y that maximizes the conditional probability of <y> given a source sentence <x>}\}.”, which is a special case of the pattern:  <noun phrase> “be equivalent to” <verb phrase>.

Example 4: “[\textit{The self-supervision task used to train BERT}] \underline{\textit{is}} \{\textit{the masked language-modeling or cloze task}\}.”, which is a special case of the pattern:  <noun phrase> “be” <noun phrase>.

Example 5: “[\textit{CLIP}] \underline{\textit{has}} \{\textit{two branches, CLIP-I and CLIP-T which encode image and text, respectively, into a global feature representation}\}.”, which is a special case of the pattern:  <noun phrase> “have” <noun phrase>.

There are two advantages of expressing definitions through subclass relation, synonymy and part-whole relation. First, the completeness of identifiers can be guaranteed because except causality, similarity, location relation, and others that are not able to define a noun phrase, the three kinds of relation cover all relations that regulate the meaning of a noun phrase. Second, the patterns are interpretable and comprehensible because different kinds of identifiers represent meanings with different patterns.

The pattern has 91.4 precision and 96.2 recall on our self-built dataset.

\subsubsection{Identifier for Subclass Relation}
<subclass identifier> ::= [“which” | “that”] “be” (“a” | “an”) [“kind of”]

Two subordinators of relative clause “that” and “which” handle the case where the definiens appears in relative clause of the definiendum, vice verse (e.g., “[\textit{Instagram}], \underline{\textit{which is a}} \{\textit{popular photo and video sharing social networking service}\} are shared on …”).

The words “be” and “a” or “an” are used to restrict a class, for example: “[\textit{Resource Space Model}] \underline{\textit{is a kind of}} \{\textit{data model which can effectively and flexibly manage the digital resources in cyber-physical system from multidimensional and hierarchical perspectives}\}.”, therefore a pattern is generated as:  <noun phrase> “be” (“a” | “an”) “kind of” <noun phrase>.

The strategy of matching the pattern of identifier and words within sentence is:  Choose the longest one if two or more identifiers are de-tected (e.g., “be a kind of” is selected if “be”, “be a” and “be a kind of” are matched).

\subsubsection{Identifier for Part-Whole Relation}
<part-whole identifier> ::= [“which” | “that”] (”have” | "consist of” | "make up") | ([“be”] “include" (“in” | ”to” | ”into”)) | ([“be”] “make up" (“of” | ”from”)) | ([“be”] (“decompose" | “add") (“in” | ”to” | ”into” | ”between” | “among”))

Example 6, “[\textit{The first form of data augmentation}] \underline{\textit{consists of}} \{\textit{generating image translations and horizontal reflections}\}.”, which is a special case of the pattern: <noun phrase> “consist of” <verb phrase>. The phrase “consists of” is used to list two parts “generating image translations” and “generating horizontal reflections” of “The first form of data augmentation”.

Passive is often used in language as described in example 7: “… \{\textit{any delays in the TOGGLE and XOR modules}\} \underline{\textit{are included in}} [\textit{the overall control signal delay}]“.  To reflect this case, the following pattern is generalized: <noun phrase> “be include in” <noun phrase>.

Example 8: “\{\textit{Yes/No questions}\} \underline{\textit{make up}} [\textit{a subset of the reading comprehension datasets CoQA, QuAC, and HotPot-QA}] …”.  It can be generalized as: <noun phrase> “make up” <noun phrase>. The verb “make up” indicates that the subject is parts of the object as the whole.

Example 9: “[\textit{The objective function used in our optimization}] \underline{\textit{is composed of}} \{\textit{a contrastive loss and a softmax cross entropy}\}.”, which can be generalized as: <noun phrase> “be compose of” <noun phrase>.

Example 10: “[\textit{The centered estimator formula}] \underline{\textit{can be decomposed into}} \{\textit{the sum of the uncentered estimator formula and the term formula}\}.”, which can be generalized as: <noun phrase> “be decompose into” <noun phrase>. The verb “decompose” indicates the meaning of splitting a whole into parts.

Example 11: “\{\textit{Variational dropout}\} \underline{\textit{is added to}} [\textit{the input of both biLSTM layers}].” can be generalized as: <noun phrase> “be add to” <noun phrase>. The verb “add” indicates the meaning of putting a thing into another thing to complement the whole.

According to the above discussion, the following patterns can be generalized:

<noun phrase> [“which” | “that”] (“have” | "consist of") | ([“be”] (“make up" (“of” | ”from”)) | (“decompose" (“in” | ”to” | ”into” | ”between” | “among”))) (<noun phrase> | <participle phrase> | <nominal clause>)

and

(<noun phrase> | <participle phrase> | <nominal clause>) [“which” | “that”] (“make up" | ([“be”] (“include" (“in” | ”to” | ”into”)) | (“add" (“in” | ”to” | ”into” | ”between” | “among”))) <noun phrase>

\subsubsection{Identifier for Synonymy}
Synonymy is used to define a concept by another equal concept. A concept can also be represented by a verb phrase or a main clause. In example 3, the phrase “\textit{is equivalent to}” is used to describe that “\textit{finding a target sentence y that …}” is an equal concept of “\textit{translation}”. This way of definition can be generalized as: <noun phrase> [“be”] “equivalent to” <participle phrase>.

<synonymy identifier> ::= [“which” | “that”] ((“be” | "denote") [“to” | “that”]) | “namely” | “become” | (“call” | “define” | “introduce” | “view”) | ([“be”] “equivalent to”) | ([“be”] “call” [“as”]) | ([“be”] (“represent" | “define”) (“by” | “as”)) | ([“be”] (”introduce”|  “view”) “as”)

Example 12: “[\textit{The initial idea of the BTM}] \underline{\textit{is to}} \{\textit{use the word pairs generated in the whole corpus to learn the topic of short texts and reduce the dimension …}\}”, which can be generalized as: <noun phrase> “be to” <verb phrase>.

Example 13: “[\textit{The simplest linked data protocol}] \underline{\textit{is that}} \{\textit{for a given graph G, dereferencing the URI of any node x in G will return all arcs in and out of that node}\}.”, which can be generalized as: <noun phrase> “be that” <main clause>.

Example 14: “[\textit{CBOW}] \underline{\textit{denotes}} \{\textit{the vectors available on the word2vec website that are trained with word and phrase vectors on 100B words of news data}\}.”, which can be generalized as: <noun phrase> “denote” <noun phrase>.

Example 15: “[\textit{The inter-dependencies\textit{} among EDUs}] \underline{\textit{are represented by}} \{\textit{conventional rhetorical relations, e.g., Elaboration, Span, Condition, Attribution}\}.”, which can be generalized as: <noun phrase> “be represent by” <noun phrase>. The phrase “are represented by”, which is the passive of “represent”, can also be used to express synonymy.

Example 16: “[\textit{The row vectors vi in VT}] \underline{\textit{are called}} \{\textit{principal components basis vectors}\}.”, which can be generalized as: <noun phrase> “be call” <noun phrase>. The phrase “\textit{are called}” is used to give a concept another name. The following pattern can be generalized: <noun phrase> ((“be” (“call” | “name”) [“as”]) | ”namely”) <noun phrase>.

Example 17: “[\textit{A (joint) shift-reduce parser}] \underline{\textit{is defined by}} \{\textit{a distribution formula over next moves m given the top and next-to-top stack labels s1 and s2}\}.”, which can be generalized as: <noun phrase> “be define by” <noun phrase>. 

Example 18: “[\textit{Adaptive mixtures of experts (ME) and hierarchical mixtures of experts (HME)}] \underline{\textit{have been introduced as}} \{\textit{a divide and conquer approach to supervised learning in static connectionist models}\}.”, which can be generalized as: <noun phrase> “be introduce as” <noun phrase>. The word “\textit{introduce}”, with the meaning of bringing something new into the article, indicates that there is usually a definition following the proposed concept.

Example 19: “[\textit{Dynamic routing}] \underline{\textit{can be viewed as}} \{\textit{a parallel attention mechanism that allows each capsule at one level to attend to some active capsules at the level below and to ignore others}\}.”, which can be generalized as: <noun phrase> “be view as” <noun phrase>.

Example 20: “We \underline{\textit{call}} [\textit{computational graph or flow graph}] \{\textit{the graph that relates inputs and parameters to outputs and training criterion}\}.”, which can be generalized as: “call” <noun phrase> <noun phrase>, the active of “\textit{be call}”.

Example 21: “In this paper we \underline{\textit{describe}} [\textit{an image interpretation system}] \{\textit{which combines segmentation and recognition into the same inference process}\}”, which can be generalized as: “describe” <noun phrase> <relative clause>.

According to the above discussion, the following pattern can be generalized as:

<noun phrase> [“which” | “that”] (“be” (“to” | “that”)) | ("denote" [“to” | “that”]) | “namely” | “become” | ([“be”] (“equivalent to”) | (“call” [“as”]) | ((“denote" | “define”) (“by” | “as”)) | ((”introduce”|  “view”) “as”)) (<noun phrase> | <nominal clause> | <verb phrase> | <main clause>)

and

(”introduce” | “define” | “view” | “call”)  <noun phrase> (<noun phrase> | <relative clause>)

\section{Extracting Method Pattern Within Sentence}
Here focuses on the method that is represented by a sentence that describes the measure to achieve a purpose.

\subsection{Basic Method for Discovering Method Pattern}
The basic method consists of the following steps:

1.	\textit{Use a small set of examples of method} within selected papers to generalize initial patterns of the method according to basic syntax patterns, e.g., <subject> <verb> <object> <adverbial clause>. Compared to the method for identifying definition, this step incorporates adverbial clause to represent purpose and measure.

2.	\textit{Find the part-of-speech of the subject, object and adverbial clause} to specialize the patterns, e.g., “\textit{We go home by riding bike}” which is a case of <noun> <verb> <noun> “by” <verb> <noun> where “by” <verb> <noun> represents an adverbial clause.   Compared to the method for identifying definition, this step incorporates adverbial to represent purpose and measure.

3.	Generalize pattern by \textit{phrase and clause according to the result of part-of-speech}, e.g., pattern <noun> <verb> <noun> “by” <verb> <noun> can be generalized as <noun phrase> <verb phrase> “by” <verb phrase>, where <verb phrase> includes at least a verb and an optional noun phrase, and it can be further generalized as <main clause> “by” <verb phrase>, where <main clause> includes at least a noun phrase and a verb phrase. This step corresponds to step 3 of the method for definition.

4.	Use more examples to \textit{find a different abstraction} on the representation of the subject, object and adverbial clause, e.g., <noun phrase> “require” <noun phrase>, <verb phrase> “be a way to” <verb phrase> and <main clause> “if” <main clause>, therefore forming a more general pattern: (<noun phrase> “require” <noun phrase>) | (<verb phrase> “be a way to” <verb phrase>) | (<main clause> “by” <verb phrase>) | (<main clause> “if” <main clause>).  Compared to the method for identifying definition, this step incorporates adverbial to represent purpose and measure.

5.	\textit{Generalize the verb, noun, and adjective} by lowercasing and lemmatizing them as identifier.  This step corresponds to step 5 of the method for definition.

6.	Same as the step 6 of the method for definition.

7.	Same as the step 7 of the method for definition.

8.	Same as the step 8 of the method for definition.

\subsection{The General Pattern}
\subsubsection{High-Level Pattern}
The method sentence is usually identified by the phrase representing measure (in short MEA, e.g., “by”, “require”, “be dependent on”) and purpose (in short PUR, e.g., “to”, “be beneficial for”, “make it easy to”, “give an opportunity to”). Both measure and purpose can be a noun phrase, a verb phrase or a main clause.

The high-level pattern of method can be generalized as follows:

<method> ::= ((<noun phrase> | <participle phrase> | <main clause>) <identifier> (<noun phrase> | <verb phrase> | <main clause>)) |
(<identifier> (<noun phrase> | <verb phrase> | <main clause>) (<noun phrase> | <verb phrase> | <main clause>))

The pattern has 90.1 precision and 97.2 recall on our self-built dataset.

Identifier is a word or phrase that indicates the definition of a method within a sentence. Measures are originated from purpose, so humans are used to representing measure together with purpose within one sentence.  Therefore, it is reasonable to find measure by identifying the representation of its purpose or vice versa. Thus, the pattern of identifier is as follows:

<identifier> ::= <purpose identifier> | <measure identifier>

The complete pattern is presented in Appendix A.

The following are examples of representing method in a sentence.

Example 22: “[\textit{We can use the forward algorithm}] \underline{\textit{to}} \{\textit{efficiently compute the generation probability assigned to a document by a content model}\}.”, where the content within "[ ]" marks the measure, and the content within "\{ \}" marks the purpose, the content underlined "\_" marks the identifier. This indicates a special pattern of method: <main clause> “to” <verb phrase>, where the word “\textit{to}” leads an adverbial clause of purpose.

The adverbial clause can appear in front or back of the main clause.

Example 23: “\underline{\textit{For}} \{\textit{training higher layers of weights}\}, [\textit{the real-valued activities of the visible units in the RBM were the activation probabilities of the hidden units in the lower-level RBM}].”, where the main clause within “[ ]” represents a method for the purpose represented by the purpose clause led by “\textit{for}”. This example can be gseneralized as ”for” <verb phrase> <main clause>.

\subsubsection{Identifier for purpose}
The identifier can be classified into two types:

(1)	\textit{Subordinators}: Subordinators of adverbial clause of purpose (i.e., subordinator for purpose) lead the adverbial clause of purpose for linking the purpose to the measure represented by the main clause.   The widely used set of subordinators consists of “for”, “to”, “in order to”, “in order for”, “so as to”, “so as for”, “so that”, “such that”, “in order that” and “so as that”. To represent all subordinators for purpose, the following pattern is generalized according to the above discussion: (<main clause> <subordinator for purpose> <verb phrase>) | (<subordinator for purpose> (<noun phrase> | <verb phrase>) <main clause>) where <subordinator for purpose> ::= (["in order" | "so as"] ("to" | "for")) | (("so" | "such" | "in order" | "so as") "that").

(2)	\textit{Verbs and phrases that indicate the purpose of the measure represented by the subject or main clause}, e.g., “[\textit{out-of-format training}] \underline{\textit{can indeed help}} \{\textit{a QA model in nearly every case}\}”, where the verb “help” indicates the purpose (described within “\{ \}”) of the measure (described within “[ ]”). The pattern is as follows:
 (<noun phrase> | <participle phrase> | <main clause>) <others for purpose> (<noun phrase> | <verb phrase>)
 
<others for purpose> ::= [“which” | “that”] ([“be”] (“capable” (“to” | ”for” | ”in” | “of”)) | ((“use” | “require”) (“to” | “for” | “in” | “at” | “on” | “between” | “among”))) | ("open" [<premodifier>] "opportunity" ("in" | "to" | "for")) | (“make” <noun phrase> "easy" (“to” | “for” | “in”)) | (("boost" | "attempt") [“at” | ”in” | “to” | “for”]).

The following are examples that explain the patterns mentioned above.

Example 24: “[\textit{TEXTRUNNER}] \underline{\textit{is capable of}} \{\textit{responding to queries over millions of tuples at interactive speeds due to a inverted index distributed over a pool of machines}\}.”, which indicates a special pattern: <noun phrase> “be capable of” <verb phrase>, where the adjective “\textit{capable}” describes the ability of the measure represented by the subject that helps to achieve the purpose.

Example 25: “[\textit{Our protocol}] \underline{\textit{can be used in}} \{\textit{the presence of any learning model, including those that acquire additional statistical constraints from observed data while learning}\}”, which indicates a special pattern: <noun phrase> “be use in” <noun phrase>.

Example 26: “[\textit{Our approach}] \underline{\textit{}\textit{opens an opportunity to}} \{\textit{better answer long-standing questions in social science about the dynamics of information}\}.”, which can be generalized as: <noun phrase> “open an opportunity to” <verb phrase>.

Example 27: “[\textit{Natural language inference}] \underline{\textit{makes it easy to}} \{\textit{judge the degree to which neural network models for sentence understanding capture the full meanings for natural language sentences}\}.”, which can be generalized as: <noun phrase> “make it easy to” <verb phrase>.

Example 28: “[\textit{neural machine translation}] \underline{\textit{attempts to}} \{\textit{build and train a single, large neural network that reads a sentence and outputs a correct translation}\}.”, which can be generalized as: <noun phrase> “attempt to” <verb phrase>.

Example 29: “[\textit{A heavily parameterized large pretrained encoder}] \underline{\textit{can boost}} \{\textit{knowledge transfer to resourcepoor languages}\}.”, which can be generalized as: <noun phrase> “boost” <noun phrase>. 

The following is an example where the measure is represented by a main clause (described within “[ ]”).

Example 30: “[\textit{A larger GIF dictionary will also improve the reaction similarity’s accuracy}], \underline{\textit{offering new approaches for}} \{\textit{studying relationships between reactions}\}.”, which is a special case of pattern: <main clause> ("open" <premodifier> "opportunity" ("in" | "to" | "for")) <verb phrase>.

\subsubsection{Identifier for measure}
The identifier can be classified into two types:
(1)	\textit{Subordinators}: Subordinators of adverbial clause of manner (i.e., "by", "via", "through", ”with”, "benefit from", "by means of" and “come from the use of”), condition (i.e., "if", “only if”, "if and only if", “on the basis of” and “on condition that”) and time (i.e., “after” and “once”) lead these adverbial clauses for linking the measure to the purpose represented by the main clause.   To represent all subordinators for measure, the following pattern is generalized according to the above discussion:

(<main clause> <subordinator for measure> <verb phrase>) | (<subordinator for measure> (<noun phrase> | <verb phrase>) <main clause>)

<subordinator for measure> ::= ("by" | "via" | "through" | ”with” | "benefit from" | "by means of" | “come from the use of”) | ("if" | "if and only if" | “on the basis of” | “on condition that”) | (“after” | “once”).

The following are examples that explain the patterns mentioned above.

Example 31: “\underline{\textit{By}} [\textit{introducing such corruptions in OOD datasets}], \{\textit{the calculated mean detection error for both CIFAR-10 and CIFAR-100 is 0\%}\}.”, which indicates a special pattern: “by” <verb phrase> <main clause>, where the preposition “by” leads an adverbial clause of manner that represents the measure for the purpose.

Measure can be viewed as a necessary or sufficient condition of achieving purpose so the words or phrases representing condition can be the identifier of finding measure and purpose.

Example 32: “\{\textit{A prediction is correct}\} \underline{\textit{if and only if}}\textit{} [\textit{all its predicted sentiment elements in the pair or triplet are correct}].”, where the adverbial clause appears after the main clause.

(2)	\textit{Verbs and phrases that indicate the measure of the purpose represented by the subject or main clause}.  The pattern is as follows:

((<noun phrase> | <participle phrase> | <main clause>) <others for measure> (<noun phrase> | <verb phrase>)) | (<others for measure> (<noun phrase> | <verb phrase>) <main clause>)

 <others for measure> ::= [“which” | “that”] (“require” [“to” | for]) | ([“be”] “rely” ("in" | "on" | "upon" | "around"))
 
Example 33: “\{\textit{Human-level AI}\} \underline{\textit{also requires}}\textit{} [\textit{the ability to transcend the outermost context the system has used so far}].”, which indicates a special pattern: <noun phrase> “require” <noun phrase>, where the verb “require” indicates a condition.

Example 34: “However, \{\textit{the effect of batch normalization}\} \underline{\textit{is dependent on}}\textit{} [\textit{the minibatch size}] …”, which can be generalized as: <noun phrase> “be dependent on” <noun phrase>.

The following is an example where the purpose is represented by a main clause (described within “{ }”).

Example 35: “\underline{\textit{Depending on}} [\textit{the design of the reader model}], \{\textit{these systems could be further categorised into extractive models and generative models}\}.”, which is a special case of pattern: <identifier for measure> <noun phrase> <main clause>.

\section{Integration Of Definitions And Methods}
\subsection{Integration of Definition}
The integration of definition is to integrate relevant definition sentences on the same concept.

Two definition sentences can be integrated into one definition according to the following conditions:

1.	The definiendums of two sentences have common word or phrase within the given text as the two sen-tences belong to the same definitions. For example, “[\textit{transformer}] \underline{\textit{is a}} \{\textit{permutation-invariant architecture without explicit positional encoding}\}.” and “[\textit{Transformers}] \underline{\textit{are a class of}} \{\textit{neural architectures that have made a tremendous transformative impact on modern natural language processing research and applications}\}.” can be integrated together because they have the common phrase “\textit{transformer}”.  The matching between two phrases is implemented by algorithm 2. Assume a dictionary consisting of different usages of a phrase and synonyms is available.  Using the dictionary, a function \textit{Usage} (phrase) that transforms an input phrase into a set of different forms of its usage or its synonyms can be designed.  Each phrase is transformed into a Semantic Link Network (SLN) \citep{2010The} \citep{Cyber-Physical-Social} and the similarity between two SLNs is calculated. If the similarity between two SLNs is larger than a given threshold, then two phrases is matched. The approach adopt a SLN-based similarity calculation by adapting the approach introduced in \citep{DBLP:journals/eswa/CaoZ22}, the detail is shown in Appendix D.

2.	The definiendum of one sentence and the definiens of the other sentence have common word or phrase within the given text as one sentence further explains the other.  \textit{For example}, “[\textit{the output word}] \underline{\textit{is also represented by}}\textit{} \{\textit{its feature vector}\}.” and “[\textit{The feature vector}] \underline{\textit{represents}} \{\textit{different aspects of the word: each word is associated with a point in a vector space}\}” can be integrated because “its feature vector” and “The feature vector” have the same phrase “feature vector” where “The” and “its” are stopwords.

3.	The first sentence is followed by another definition sentence where its subject is a pronoun or the premodifier of its subject contains the determiners “this”, “that”, “these”, “those” or possessive pronouns “its”, “their”. For example, “[\textit{KIF}] \underline{\textit{is a}} \{\textit{frame\textit{}work for exchanging of declarative knowledge among heterogeneous computer systems}\}” and “[\textit{It}] \underline{\textit{is a}} \{\textit{version of first order predicate calculus with extensions to support non-monotonic reasoning and quoting}\}”.

The following is an example of integrating 7 definitions:

(1) “[\textit{Neural machine translation}] \underline{\textit{is a}} \{\textit{recently proposed approach to machine translation}\}.”;

(2) “[\textit{translation}] \underline{\textit{is equivalent to}} \{\textit{finding a target sentence y that maximizes the conditional probability of y given a source sentence x}\}.”;

(3) “[\textit{This neural machine translation approach}] \underline{\textit{typically consists of}} \{\textit{two components, the first of which encodes a source sentence formula and the second decodes to a target sentence formula}\}.”;

(4) “[\textit{The models proposed recently for neural machine translation}] \underline{\textit{often belong to}} \{\textit{a family of encoder-decoders}\} \textit{and encode a source sentence into a fixed-length vector from which a decoder generates a result}.”;

(5) “[\textit{The whole encoder-decoder system}], \underline{\textit{which consists of}} \{\textit{the encoder and the decoder for a language pair}\}, \textit{is jointly trained to maximize the probability of a correct translation given a source sentence}.”;

(6) “[\textit{Its decoder}] \underline{\textit{has}} \{\textit{1000 hidden units}\}.”;

(7) “[\textit{The most important distinguishing feature of this approach from the basic encoder–decoder}] \underline{\textit{is that}} \{\textit{it does not attempt to encode a whole input sentence into a single fixed-length vector}\}.”.

Sentences (1), (2), (3) and (4) can be integrated together according to rule 1 because the definiendum of (1) shares a common phrase with the definiendums of (2), (3) and (4). Sentences (4), (5) and (6) can be integrated together according to rule 2 because the definiendums of (5) and (6) share common phrase with the definiens of (4). Sentences(6) and (7) can be integrated together according to rule 1.

\subsection{Integration of Method}
Two method sentences can be integrated into one method according to the following conditions:

1.	The measures or purposes of two sentence have common word or phrase. This indicates that methods in the two sentences have the same purpose or measure, or one method further achieves the other. For example, “\{\textit{Human-level AI}\} \underline{\textit{requires}} [\textit{reasoning about strategies of action, i.e., action programs}].” and “\{\textit{Human level artificial intelligence}\} \underline{\textit{also requires}} [\textit{the ability to transcend the outermost context the system has used so far}].”

2.	The first sentence is followed by another method sentence where its subject is a pronoun or the premodifier of its subject contains the determiners “this”, “that”, “these”, “those” or possessive pronouns “its”, “their”. For example, “\underline{\textit{In order to}} \{\textit{determine the target language in such scenarios}\} [\textit{they proposed adding dedicated target-language symbols to the source}].” and “[\textit{This method}] \underline{\textit{enabled}} \{\textit{zero-shot translation, showing the ability of the model to generalize to unseen pairs}\}.”.

The following is an example of integrated method:

(1) “[\textit{The EM algorithm can be used}] \underline{\textit{to}} \{\textit{fit credibility networks to data}\}.”;

(2) ”[\textit{credibility networks}] \underline{\textit{are able to}} \{\textit{perform segmentation and recognition simultaneously , removing the need for ad-hoc segmentation heuristics}\}”;

(3) “[\textit{credibility networks}] \underline{\textit{avoid}} \{\textit{the usual problems associated with a bottom-up approach to image interpretation}\}”;

(4) “\textit{In section 5 we demonstrate that} [\textit{binary credibility networks}] \underline{\textit{are useful in}} \{\textit{solving the problem of classifying and segmenting binary handwritten digits}\}.”;

(5) “\{\textit{the EM algorithm}\} \underline{\textit{relies on}} [\textit{integrating over the distribution of formula, with the auxiliary function formula which is the expected value of the complete data log-likelihood}]”

Sentence (1) can be integrated with (2), (3) and (4) because the purpose of (1) shares a common phrase with the measures of (2), (3) and (4).   Sentences (2), (3) and (4) can be integrated together because their measures share common phrase with each other.  Sentences (1) and (5) can be integrated together because the measure of (1) shares a common phrase with the purpose of (5).

\subsection{Integrating Definition with Method}
A definition sentence can be integrated with a method sentence to complement each other if the definiendum of a definition sentence and the measure (or purpose) of a method have common word or phrase.  For example, the definition “[\textit{A convolutional neural network, or CNN}], \underline{\textit{is a}} \{\textit{deep learning neural network designed for processing structured arrays of data such as images}\}.” and method “[\textit{deep convolutional neural network}] \underline{\textit{is capable of}} \{\textit{achieving record breaking results on a highly challenging dataset using purely supervised learning}\}” can be integrated because they have the common phrase “convolutional neural network”.

The following is an example of integrating definition with method:

(1) “\textit{We propose a new simple network architecture}, \{\textit{the Transformer}\}, \underline{\textit{based solely on}} [\textit{attention mechanisms}], dispensing with recurrence and convolutions entirely.”;

(2) ”[\textit{Attention mechanisms}] \underline{\textit{have become}} \{\textit{an integral part of compelling sequence modeling and transduction models in various tasks}\}, \textit{allowing modeling of dependencies without regard to their distance in the input or output sequences}.”;

(3) “[\textit{Self-attention, sometimes called intra-attention}] \underline{\textit{is an}} \{\textit{attention mechanism relating different positions of a single sequence in order to compute a representation of the sequence}\}.”;

(4) “[\textit{Self-attention}] \underline{\textit{has been used successfully in}} \{\textit{a variety of tasks including reading comprehension, abstractive summarization, textual entailment and learning task-independent sentence representations}\}”;

Sentences (2) and (3) are integrated because the definiendum of (2) shares common phrase with the definiens of (3).  Sentences (1) and (2) are integrated because the measure of (1) and the definiendum of (2) have common phrase.  Sentences (3) and (4) are integrated because the definiendum of (3) and the measure of (4) have common phrase.

\section{Experiments}
\subsection{Experiment on Self-Built Dataset}
Our self-built dataset includes 200 scientific papers and 38512 sentences in the computer and science field including Natural Language Processing, Computer Vision, Image Processing, Computer Graphics, Programming Languages, etc.

\subsubsection{Precision and Recall of Patterns}
Statistics is made on each class of definition and method in terms of the number of annotated sentences, precision, recall, and F1-score for the task of sentence classification. In this task, a sentence is a definition sentence if it contains at least a definiendum and a corresponding definiens, a sentence is a method sentence if it contains at least a measure and a corresponding purpose.

A baseline that is made up of definition patterns proposed by previous related works \citep{Liu2003MiningTC} \citep{articleAnswering} \citep{10.1145/354756.354817} \citep{Jin2013MiningST} \citep{10.1145/354756.354817} is also tested in self-built dataset to compare with ours. The detailed patterns of the baseline are shown in Appendix E.

\begin{table}[H] \small
\centering
\begin{tabular}{lllll}
\hline
Type & Annotated & Precision & Recall & F1-Score\\
\hline
Definition & 9855 & 92.0 & 95.9 & 93.9 \\
Previous & N/A & 87.5 & 39.0 & 54.0 \\
Method & 12577 & 93.7 & 96.5 & 95.1 \\
\hline
\end{tabular}
\caption{\label{citation-guide}
Experiment for sentence classification.
}
\end{table}

Both the precision and recall of previous definition patterns are lower than our pattern.

Algorithm 1 for extracting definitions and methods is described in Appendix G.

\subsubsection{Completeness verification}
To verify the completeness of patterns, the self-built dataset is split into two parts: the patterns summarized when the first 162 papers were annotated are recorded, and they are verified on the remaining 38 papers.

\begin{table}[H] \small
\centering
\begin{tabular}{lllll}
\hline
Type & Annotated & Precision & Recall & F1-Score\\
\hline
Definition & 7823 & 90.4 & 95.2 & 92.7 \\
Method & 10215 & 93.0 & 95.5 & 94.2 \\
\hline
\end{tabular}
\caption{\label{citation-guide}
Experiment for sentence classification on test set.
}
\end{table}

The experimental results have not changed much: in sentence classification, the F1-score decline by 1.0 in definition and 0.9 in method. It means that the patterns obtained from the first 162 papers basically match the patterns proposed obtained from 200 papers. This proves the completeness of pattern.

\begin{figure}[!t]
	\centering
	\includegraphics[width=3in]{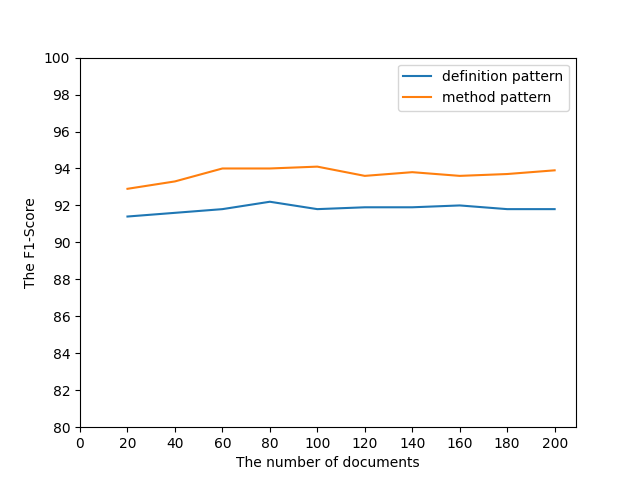}
	\caption{The F1-Score changes with the number of documents.}
	\label{fig_sim}
\end{figure}

As shown in Fig. 1, the F1-score of definition and method pattern have hardly changed as the number of documents increases.

\subsubsection{Precision and Recall of Integration Approach}
Complete definitions and methods within five papers were annotated to test the precision and recall of the integration approach. Table 3 shows the result of experiment for integrating definitions and integrating methods.

\begin{table}[H] \small
\centering
\begin{tabular}{lllll}
\hline
Type & Annotated & Precision & Recall & F1-Score\\
\hline
Definition & 49 & 71.2 & 73.0 & 72.1 \\
Method & 61 & 73.5 & 69.3 & 71.3 \\
\hline
\end{tabular}
\caption{\label{citation-guide}
Experiment for integration approach.
}
\end{table}

Some definition sentences or method sentences cannot be integrated due to that the dictionary cannot cover all equivalent phrases, e.g., “\textit{Human-level intelligence}” and “\textit{Human level artificial intelligence}”. A more complete dictionary helps improve the performance of extracting complete definitions and methods.  If this problem is solved, the result can be improved as follows: 74.0 precision and 76.2 recall for definition, 73.5 precision and 72.1 recall.

\subsection{Experiment on Public Dataset}
\subsubsection{Experiment on WCL Dataset}
The WCL dataset consists of 1717 sentences annotated as definition and 2847 sentences annotated as non-definition.

The description of each baseline used in this dataset is given:

(1)	Previous, a collection of patterns proposed by previous related works. Its details are shown in Appendix E.

(2)	Bigrams \citep{10.1145/1229179.1229182}, a bigram language model for generating the probability of a sentence belonging to definition.

(3)	WCLs \citep{10.5555/1858681.1858815}, which clusters and aligns the definition examples into directed acyclic graphs (DAG) and judges whether the test sentence can be transformed into one of subgraph.

(4)	DefMiner \citep{Jin2013MiningST}, which uses conditional random fields with 12 hand-crafted features (e.g., whether a word is in the keyphrase list of the paper, distance from a word to the root of the sentence in the dependency tree) to label the definiendum and definiens in a sentence. If both definiendum and definiens appear in a sentence, the sentence is definition.

(5)	B\&DC \citep{Boella2013ExtractingDA} and E\&S \citep{10.1007/978-3-319-07983-7_10}, which use features vectors whose each dimension represents a dependency-based feature (e.g., whether a specific topological structure appears in a dependency tree) as the input of machine learning models (e.g., support vector machine, naïve bayes).

(6)	LSTM-POS \citep{10.1007/978-3-319-47674-2_16}, which uses word vectors of each word or its part-of-speech within a sentence as the input of recurrent neural network with Long Short-Term Memory to classify the sentence.

\begin{table}[H] \small
\centering
\begin{tabular}{llll}
\hline
Models & Precision & Recall & F1-Score\\
\hline
Our Pattern & 89.6 & 94.5 & 92.0 \\
Previous & 84.1 & 75.0 & 79.3 \\
Bigrams & 66.7 & 82.7 & 73.8 \\
WCLs & 98.8 & 60.7 & 75.2 \\
B\&DC & 88.0 & 76.0 & 81.6 \\
DefMiner & 92.0 & 79.0 & 85.0 \\
E\&S & 85.9 & 85.3 & 85.4 \\
LSTM-POS & 90.4 & 92.0 & 91.2 \\
\hline
\end{tabular}
\caption{\label{citation-guide}
Experiment on WCL dataset for sentence classification.
}
\end{table}

Our pattern gets the highest F1 score 92.0 and highest recall 94.5, which is better than LSTM-POS.

The precision of our approach is influenced by the factor that our understandings of definition is different from WCLs.  The following examples were not annotated but are extracted as definition by our pattern: (1) “[\textit{aggregatibacter actinomycetemcomitans}] \underline{\textit{is a}} \{\textit{gram negative TARGET which is prevalent in subgingival plaques}\}.” and (2) “[\textit{the unique feature of TARGET}] \underline{\textit{is that}} \{\textit{anyone can create their own custom social network for a particular topic or need, catering to specific audiences}\}.”.

\subsubsection{Experiment on W00 Dataset}
The W00 dataset is an annotated dataset of definitions collected from scientific papers in the area of natural language processing, and consists of 865 sentences annotated as definition and 1,647 sentences annotated as non-definition.

The baselines chosen in this dataset are

(1)	C-BLSTM \citep{Anke2018SyntacticallyAN}, which uses the average of two word vectors whose corresponding nodes in dependency tree are adjacent as the input of convolutional and recurrent neural networks to classify the sentence.

(2)	Joint \citep{articleJoint}, which uses word vectors as the input of recurrent neural network and graph convolution neural network to classify sentences.

\begin{table}[H] \small
\centering
\begin{tabular}{llll}
\hline
Models & Precision & Recall & F1-Score\\
\hline
Our Pattern & 51.7 & 78.7 & 62.5 \\
Previous & 52.5 & 42.3 & 46.8 \\
C-BLSTM & 52.0 & 67.6 & 57.4 \\
Joint & 67.0 & 68.0 & 67.2 \\
\hline
\end{tabular}
\caption{\label{citation-guide}
Experiment on W00 dataset for sentence classification.
}
\end{table}

Comparing to C-BLSTM, the precision of ours is a bit lower, but the recall is much higher and F1-score is also higher. Comparing to Joint, our precision is lower, but recall is still higher.

\section{Related Works}
\subsection{Discovering Pattern}
Approaches to discover pattern of definition focus on forming syntactic patterns, which are determined according to lexicon, syntax and relations.  Most previous approaches to forming definition patterns are based on subclass relation and synonymy because a noun phrase can be defined by subclass relation and synonymy relation.  The pattern of subclass relation can be observed by the following steps  \citep{10.3115/992133.992154}: (1) gather a list of subclass relation pairs (e.g., ”England-country”); (2) find the commonalities among the environments where these pairs appear syntactically near one another, and hypothesize that common ones yield patterns; and, (3) once a new pattern has been identified, use it to gather more instances and go to the first step.  These approaches are interpretable, provable and understandable. They are also expandable because patterns can be easily extended through “or” operation.

Other related works focus on obtaining a computing model for extracting definitions from texts.  A model for determining the probability of a sentence belonging to definition can be trained by definition examples.  The parameters of the model is for generating the sequence of definition \citep{10.1145/1229179.1229182}.  A model for determining whether a sentence is a definition or not is built by clustering and aligning the definition examples into directed acyclic graphs (DAG) and judging whether a test sentence can be transformed into one of subgraph \citep{10.5555/1858681.1858815}. Another work summarizes a set of patterns for finding definiendum and definiens based on statistic and syntactic features \citep{Jin2013MiningST}. These approaches can only be interpreted and proved through mathematical and statistical theory, and can only be extended by changing training-data or modifying models (e.g., the topological structure of graphs, the features selection of models).

C-BLSTM \citep{Anke2018SyntacticallyAN} is a deep learning model that calculates the average of any two context vectors whose corresponding words have a head-modifier relation in dependency tree to combine syntactic information of the given sentence into context vectors. Then the vectors are applied as input of downstream neural network classifier. The model introduced in \citep{articleJoint}  further combines the semantic information (e.g., the words closer to definiendum and definiens are more important than others, and the similarity between definiendum and definiens should be higher than the other parts of the sentence) into the design of loss function. This work obtains 99.5 F1-score on WCL and 67.2 F1-score on W00, which shows that semantic information is beneficial for reflecting the features of definition. These approaches usually have better performance due to the context vectors trained on large-scale dataset that not contained in target dataset and the capability of neural networks to automatically capture features, but it is difficult to find a mathematical or linguistic explanation of the change of performance. Their extensibility lies in the selection of training-data, pretrained context vectors and various neural networks, and the design of internal structure of neural networks and loss function.

Most previous works only focused on the performance of single-sentence dataset (e.g., WCL and W00) without integrating complete definitions, except the ones focused on DEFT dataset \citep{Spala2019DEFTAC} that contains multi-sentence annotation instances. But these multi-sentence definitions are consecutive (e.g., one definition sentence is followed by another sentence whose subject is a pronoun), so these works cannot extract complete definitions being scattered within a document.

\subsection{Method Extraction}
So far, no work on discovering patterns of method has been found by searching Google Scholar, ACL Anthology, Digital Bibliography \& Library Project with the combination of “method”, “extraction”, “knowledge”, “pattern” and their synonyms.

The related works focus on obtaining a computing model for extracting methods from texts. An approach to identify methodological sentences is proposed  to categorize sentences within paper into seven kinds: Background (background knowledge), Aim (goal of research), Basis (specific other work that the presented approach is based on), Contrast (contrasting and comparison to other solutions), Other (specific other work), Textual (textual structure of the paper), and Own (own work including method, results, future work) \citep{articleTeufel}.  The Own kind sentences was further split into four kinds and integrating relevant sentences together \citep{articleKovacevic}: Task(s) that are addressed by a given method, Method(s) that are used to perform a task, Resources (like datasets) and Features that the method relies on, and its Implementation. Four Conditional Random Fields (CRF) with consitiuency parse, dependency parse, frequency and others as features were trained for extracting each kind respectively.  The approach integrated methodological sentences with the same methodological segments presented within a given paper into groups.  An approach was proposed to classify methodological sentences which are essential to solve the research problems in abstracts of scientific papers \citep{10.1007/978-3-030-43687-2_66}. This approach used context vectors as the input of neural networks to classify sentences, then it applied document frequency of words in methodological sentences and other sentences respectively to filter out non-methodological sentences. Anaphoric relation was used to integrate method sentences, i.e., any two consecutive sentences where the latter starts with “This method” are judged as two method sentences \citep{inproceedingsHoungbo}.

\section{Conclusion}
This wok focuses on discovering patterns of definition and method, and approach to integrate the definition and method. The innovations of this research lie in the following aspects:

1.	Independent and more complete definition pattern.  The learning-based approaches are for generating results (definitions) but not for generating interpretable definition patterns and the results depend on the model and training data. Most previous approaches for generating definition patterns are based on subclass relation and synonymy.  In our approach, the completeness of the definition patterns is guaranteed at three levels: (1) the semantic level completeness is guaranteed by considering all the semantic relations (synonymy, antonymy, subclass relation, part-whole relation, morphological relations) that can express the definitions (synonymy, subclass relation, part-whole relation); (2) the syntactic level completeness is guaranteed by considering the different syntactic forms of the active and passive of verbs in the definition patterns; and, (3) the lexical level completeness is guaranteed by querying synonyms of the words in identifiers.

2.	Independent method pattern.  A set of method patterns is summarized for the first time.  These patterns obtain 93.7 precision and 96.5 recall on our self-built dataset.

3.	The approach for integrating definitions and methods is proposed to extract complete definitions and methods that consist of multiple sentences scattered within document. Previous works focus on extracting definitions or methods in the form of single sentence or multiple consecutive sentences, which is unable to extract complete definitions and methods.

4.	The approaches for extracting definition patterns and method pattern, which interpret the processes of extracting patterns.  The approach can be used and extended for different purposes.

Experiments on using the patterns of definition show that, our approach has much better performance than previous pattern-based approaches especially on recall, and is also comparable to learning-based approaches that lack interpretability and cannot be extended independently of their models and training datasets.

\section*{Acknowledgements}

This work was supported by the National Science Foundation of China (project no. 61876048).

\bibliography{anthology,custom}

\begin{thebibliography}{23}
\expandafter\ifx\csname natexlab\endcsname\relax\def\natexlab#1{#1}\fi

\bibitem[{Boella and Caro(2013)}]{Boella2013ExtractingDA}
G.~Boella and L.~Di Caro. 2013.
\newblock Extracting definitions and hypernym relations relying on syntactic
  dependencies and support vector machines.
\newblock In \emph{Proceedings of ACL}, page 532–537.

\bibitem[{Cao and Zhuge(2022)}]{DBLP:journals/eswa/CaoZ22}
M.~Cao and H.~Zhuge. 2022.
\newblock \href {https://doi.org/10.1016/j.eswa.2022.117777} {Automatic
  evaluation of summary on fidelity, conciseness and coherence for text
  summarization based on semantic link network}.
\newblock \emph{Expert Systems with Applications}, 206:21.

\bibitem[{Carter and McCarthy(2006)}]{Cambridgegrammar}
R.~Carter and M.~McCarthy. 2006.
\newblock \emph{Cambridge grammar of English: a comprehensive guide; spoken and
  written English grammar and usage,}.
\newblock Cambridge University Press.

\bibitem[{Chomsky(2014)}]{Theminimalistprogram}
N.~Chomsky. 2014.
\newblock \emph{The minimalist program}.
\newblock MIT press.

\bibitem[{Cui et~al.(2007)Cui, Kan, and Chua}]{10.1145/1229179.1229182}
H.~Cui, M.-Y. Kan, and T.-S. Chua. 2007.
\newblock \href {https://doi.org/10.1145/1229179.1229182} {Soft pattern
  matching models for definitional question answering}.
\newblock \emph{ACM Transactions on Information Systems}, 25(2):8–es.

\bibitem[{Espinosa-Anke and Saggion(2014)}]{10.1007/978-3-319-07983-7_10}
L.~Espinosa-Anke and H.~Saggion. 2014.
\newblock Applying dependency relations to definition extraction.
\newblock In \emph{Proceedings of NLDB}, page 63–74.

\bibitem[{Espinosa-Anke and Schockaert(2018)}]{Anke2018SyntacticallyAN}
L.~Espinosa-Anke and S.~Schockaert. 2018.
\newblock Syntactically aware neural architectures for definition extraction.
\newblock In \emph{Proceedings of HLT-NAACL}, pages 378--385.

\bibitem[{Hearst(1992)}]{10.3115/992133.992154}
M.~A. Hearst. 1992.
\newblock \href {https://doi.org/10.3115/992133.992154} {Automatic acquisition
  of hyponyms from large text corpora}.
\newblock In \emph{Proceedings of COLING}, page 539–545.

\bibitem[{Hildebrandt et~al.(2004)Hildebrandt, Katz, and
  Lin}]{articleAnswering}
W.~Hildebrandt, B.~Katz, and J.~Lin. 2004.
\newblock Answering definition questions using multiple knowledge sources.
\newblock In \emph{Proceedings of HLT-NAACL}, pages 49--56.

\bibitem[{Houngbo and Mercer(2012)}]{inproceedingsHoungbo}
H.~Houngbo and R.~E. Mercer. 2012.
\newblock Method mention extraction from scientific research papers.
\newblock In \emph{Proceedings of COLING}, page 1211–1222.

\bibitem[{Jin et~al.(2013)Jin, Kan, Ng, and He}]{Jin2013MiningST}
Y.~Jin, M.-Y. Kan, J.-P. Ng, and X.~He. 2013.
\newblock Mining scientific terms and their definitions: A study of the acl
  anthology.
\newblock In \emph{Proceedings of EMNLP}, page 780–790.

\bibitem[{Joho and Sanderson(2000)}]{10.1145/354756.354817}
H.~Joho and M.~Sanderson. 2000.
\newblock \href {https://doi.org/10.1145/354756.354817} {Retrieving descriptive
  phrases from large amounts of free text}.
\newblock In \emph{Proceedings of CIKM}, page 180–186.

\bibitem[{Kovačević et~al.(2012)Kovačević, Konjovic, Milosavljević, and
  Nenadic}]{articleKovacevic}
A.~Kovačević, Z.~Konjovic, B.~Milosavljević, and G.~Nenadic. 2012.
\newblock Mining methodologies from nlp publications: A case study in automatic
  terminology recognition.
\newblock \emph{Computer Speech and Language}, 26:105--126.

\bibitem[{Li et~al.(2016)Li, Xu, and Chung}]{10.1007/978-3-319-47674-2_16}
S.~Li, B.~Xu, and T.~L. Chung. 2016.
\newblock Definition extraction with lstm recurrent neural networks.
\newblock In \emph{Proceedings of CCL}, pages 177--189.

\bibitem[{Liu et~al.(2003)Liu, Chin, and Ng}]{Liu2003MiningTC}
B.~Liu, C.~W. Chin, and H.~T. Ng. 2003.
\newblock Mining topic-specific concepts and definitions on the web.
\newblock In \emph{Proceedings of WWW}, pages 251--260.

\bibitem[{Miller et~al.(2022)Miller, Beckwith, Fellbaum, Gross, and
  Miller}]{0Introduction}
G.~A. Miller, R.~Beckwith, C.~D. Fellbaum, D.~Gross, and K.~J. Miller. 2022.
\newblock Introduction to wordnet: An on-line lexical database.
\newblock \emph{International Journal of Lexicography}.

\bibitem[{Navigli and Velardi(2010)}]{10.5555/1858681.1858815}
R.~Navigli and P.~Velardi. 2010.
\newblock Learning word-class lattices for definition and hypernym extraction.
\newblock In \emph{Proceedings of ACL}, page 1318–1327.

\bibitem[{Spala et~al.(2019)Spala, Miller, Yang, Dernoncourt, and
  Dockhorn}]{Spala2019DEFTAC}
S.~Spala, N.~A. Miller, Y.~Yang, F.~Dernoncourt, and C.~Dockhorn. 2019.
\newblock Deft: A corpus for definition extraction in free- and semi-structured
  text.
\newblock In \emph{Proceedings of the 13th Linguistic Annotation Workshop},
  pages 124--131.

\bibitem[{Teufel and Moens(2001)}]{articleTeufel}
S.~Teufel and M.~Moens. 2001.
\newblock Summarizing scientific articles - experiments with relevance and
  rhetorical status.
\newblock \emph{Computational Linguistics}, 28:409--445.

\bibitem[{Veyseh et~al.(2020)Veyseh, Dernoncourt, Dou, and
  Nguyen}]{articleJoint}
A.~P.~B. Veyseh, F.~Dernoncourt, D.~Dou, and T.~H. Nguyen. 2020.
\newblock A joint model for definition extraction with syntactic connection and
  semantic consistency.
\newblock In \emph{Proceedings of AAAI}, page 7166–7174.

\bibitem[{Wang et~al.(2020)Wang, Zhang, Zhang, and
  Zhang}]{10.1007/978-3-030-43687-2_66}
R.~Wang, C.~Zhang, Y.~Zhang, and J.~Zhang. 2020.
\newblock Extracting methodological sentences from unstructured abstracts of
  academic articles.
\newblock In \emph{Proceedings of Sustainable Digital Communities}, pages
  790--798.

\bibitem[{Zhuge(2012)}]{2010The}
H.~Zhuge. 2012.
\newblock \emph{The Knowledge Grid: Toward Cyber-Physical Society}.
\newblock World Scientific.

\bibitem[{Zhuge(2020)}]{Cyber-Physical-Social}
H.~Zhuge. 2020.
\newblock \emph{Cyber-Physical-Social Semantic Link Network, Chapter 3 in
  Cyber-Physical-Social Intelligence on Human-Machine-Nature Symbiosis}.
\newblock Springer.

\end{thebibliography}
\bibliographystyle{acl_natbib}

\appendix

\section{Definition Pattern}
\label{sec:appendix}

\subsection{High-Level Definition Pattern}
The following are high-level patterns.

<definition> ::= (<noun phrase> (“be” | “have”) <noun phrase>) | 
(<noun phrase> <other identifier> (<noun phrase> | <nominal clause> | <verb phrase> | <main clause>)) |
((<noun phrase> | <nominal clause> | <participle phrase>) <other identifier> <noun phrase>) |
(<other identifier> <noun phrase> (<relative clause> | <noun phrase>))

The following are the definition of other identifiers.

<other identifier> ::= <subclass identifier> | <part-whole identifier> | <synonymy identifier>

<subclass identifier> ::= [“which” | “that”] "be" (“a” | “an”) [(“kind” | <synonym of kind>) “of”]

<part-whole identifier> ::= [“which” | “that”] (”have” | "consist of” | <synonym of consist of> | "make up" | <synonym of make up>) | ([“be”] (“consist" | <synonym of consist of>) (“in” | ”to” | ”into”)) | ([“be”] (“make up" | <synonym of make up>) (“of” | ”from”)) | ([“be”] (“decompose" | <synonym of decompose> | “add" | <synonym of add>) (“in” | ”to” | ”into” | ”between” | “among”))

<synonymy identifier> ::= [“which” | “that”] ((“be” | "denote" | <synonym of denote>) [“to” | “that”]) | “namely” | “become” | (“call” | “name” | “define” | <synonym of define> | “introduce” | <synonym of introduce> | “view” | <synonym of view>) | ([“be”] (“equivalent” | <synonym of equivalent>) “to”) | ([“be”] “call” | “name” [“as”]) | ([“be”] ("denote" | <synonym of denote> | “define” | <synonym of define>) (“by” | “as”)) | ([“be”] (”introduce” | <synonym of introduce> |  “view” | <synonym of view>) “as”)

The following are synonyms that are used to expand patterns by searching the WordNet.

<synonym of kind> ::= “sort” | “type” | “class”

<synonym of consist of> ::= "contain" | "involve" | "include" | "comprise" | "encompass"

<synonym of make up> ::= "compose" | "constitute" | “belong to”

<synonym of decompose> ::= "split" | "break" | "divide" | "separate" | "categorize" | "classify"

<synonym of add> ::= "supplement" | ”group” | ”combine”

<synonym of split> ::= "decompose" | "break" | "divide" | "separate" | "categorize" | "classify"

<synonym of equivalent> ::= "equal" | ”correspond” | “amount” | ”refer”
<synonym of denote> ::= "represent" | ”indicate” | ”mean”

<synonym of define> ::= ”redefine” | ”express” | “describe”

<synonym of introduce> ::= ”propose” | ”report” | ”develop” | ”design” | ”articulate”

<synonym of view> ::= ”employ” | “use” | “reuse” | “utilize” | “think” | “see” | ”regard” | “consider” | “count” | “formalize” | “select”

\subsection{Full Definition Pattern}
The following are full definition pattern identified by subclass relation:

<noun phrase> [“which” | “that”] "be" (“a” | “an”) [(“kind” | <synonym of kind>) “of”] <noun phrase>

The following are full definition pattern identified by part-whole relation:
<noun phrase> “have” <noun phrase> | 
<noun phrase> [“which” | “that”] ("consist of” | <synonym of consist of>) | ([“be”] (“make up" | <synonym of make up>) (“of” | ”from”)) | ([“be”] (“decompose" | <synonym of decompose>) (“in” | ”to” | ”into” | ”between” | “among”)) (<noun phrase> | <participle phrase> | <nominal clause>) | 
(<noun phrase> | <participle phrase> | <nominal clause>) [“which” | “that”] (“make up" | <synonym of make up>) | ([“be”] (“consist" | <synonym of consist>) (“in” | ”to” | ”into”)) | ([“be”] (“add" | <synonym of add>) (“in” | ”to” | ”into” | ”between” | “among”)) <noun phrase>

The following are full definition pattern identified by synonymy:

<noun phrase> [“which” | “that”] ((“be” | "denote" | <synonym of denote>) [“to” | “that”]) | “namely” | “become” | ([“be”] (“equivalent” | <synonym of equivalent>) “to”) | ([“be”] “call” | “name” [“as”]) | ([“be”] ("denote" | <synonym of denote> | “define” | <synonym of define>) (“by” | “as”)) | ([“be”] (”introduce” | <synonym of introduce> |  “view” | <synonym of view>) “as”) (<noun phrase> | <nominal clause> | <verb phrase> | <main clause>) | 
(“call” | “name” | “define” | <synonym of define> | “introduce” | <synonym of introduce> | “view” | <synonym of view>) <noun phrase> (<noun phrase> | <relative clause>)

\section{Method Pattern}
\label{sec:appendix}

\subsection{High-Level Method Pattern and Identifier}
The following are high-level method patterns.

<method> ::= ((<noun phrase> | <participle phrase> | <main clause>) <identifier> (<noun phrase> | <verb phrase> | <main clause>)) | 
((<noun phrase> | <participle phrase> | <main clause>) <identifier> (<noun phrase> | <verb phrase> | <main clause>)) | 
(<identifier> (<noun phrase> | <verb phrase> | <main clause>) (<noun phrase> | <verb phrase> | <main clause>))

<identifier> ::= <purpose identifier> | <measure identifier>

<purpose identifier> ::= <subordinator for purpose> | <others for purpose>

<subordinator for purpose> ::= (["in order" | "so as"] ("to" | "for")) | (("so" | "such" | "in order" | "so as") "that")

<others for purpose> ::= [“which” | “that”] ([“be”] (“capable” | <synonym of capable>) (“to” | ”for” | ”in” | “at”| “of”)) | ([“be”] (“use” | <synonym of use> | “require” | <synonym of require>) (“to” | “for” | “in” | “at” | “on” | “between” | “among”)) | (("open" | <synonym of open>) [<premodifier>] ("opportunity" | <synonym of opportunity>) [<postmodifier>] ("in" | "to" | "for")) | (“make” <noun phrase> ("easy" | <synonym of easy>) (“to” | “for” | “in”)) | (("boost"  | <synonym of boost> | "attempt" | <synonym of attempt>) [“at” | ”in” | “to” | “for”])

<measure identifier> ::= <subordinator for measure> | <others for measure>

<subordinator for measure> ::= ("by" | "via" | "through" | ”with” | "benefit from" | "by means of" | “come from the use of”) | ("if" | "if and only if" | “on the basis of” | “on condition that”) | (“after” | “once”)

<others for measure> ::= [“which” | “that”] ((“require” | <synonym of require>) [“to” | ”for”]) | ([“be”] (“rely” | <synonym of rely>) ("in" | "on" | "upon" | "around"))

The following are synonyms that are used to expand patterns by searching WordNet.

<synonym of capable> ::= "useful" | "important" | "significant" | "informative" | "critical" | "beneficial" | "effective" | "sufficient" | "necessary" | "ideal" | "vital" | "essential" | "key" | "central" | "applicable" | "able" | "robust" | "good" | "advantageous" | "suit" | "suitable" | "responsible" | "capable" | "able" | "desirable"

<synonym of use> ::=  “reuse” | “utilize” | “employ”

<synonym of open> ::= "play" | "retain" | "give" | "offer" | "provide" | "introduce" | "present" | "suggest"

<synonym of opportunity> ::= "chance" | "freedom" | "method" | "way" | "avenue" | "approach" | "tool" | "door" | "role"

<synonym of easy> ::= "possible" | "viable" | "effective" | "sufficient" | "convenient" | "practical"

<synonym of boost> ::= "permit" | "help" | "enable" | "ensure" | "make" | "avoid" | "facilitate" | "improve" | "achieve" | "reach" | "yield" | "alleviate" | "solve" | "address" | "handle" | "tackle" | "grapple"

<synonym of attempt> ::= "aim" | "focus" | "concentrate"

<synonym of require> ::= "need" | "demand" | "necessitate"

<synonym of rely> ::= "depend" | "base" | "build" | "rely" | “dependent”

\subsection{Full Method Pattern}

The following are full method pattern identified by purpose:

(<main clause> <subordinator for purpose> <verb phrase>) | (<subordinator for purpose> (<noun phrase> | <verb phrase>) <main clause>) | 
(<noun phrase> | <participle phrase> | <main clause>) <others for purpose> (<noun phrase> | <verb phrase>)

The following are full method pattern identified by measure:

(<main clause> <subordinator for measure> <verb phrase>) | (<subordinator for measure> (<noun phrase> | <verb phrase>) <main clause>) | 
((<noun phrase> | <participle phrase> | <main clause>) <others for measure> (<noun phrase> | <verb phrase>)) | (<others for measure> (<noun phrase> | <verb phrase>) <main clause>)

\section{Syntax Pattern}
\label{sec:appendix}

The patterns of clause, phrase and words are summarized according to linguistic literature \citep{Theminimalistprogram} \citep{Cambridgegrammar}.

\subsection{Clause-Level Pattern}

<main clause> ::= (<noun phrase> | <person> | <nominal clause> | <participle phrase>) <verb phrase> \{<adverbial clause>\} \{[<coordinator>] <main clause>\}

<adverbial clause> ::= [<adverbial subordinator>] [<noun phrase> | <person>] <verb phrase> \{[<coordinator>] <adverbial clause>\}

<adverbial subordinator> ::= “after” | ”although” | ”as” | ”because” | ”before” | ”for” | ”how” | ”however” | ”if” | ”in case” | ((”in order” | ”so as”) (“that” | ”to” | ”for”)) | ”lest” | ”once” | ”since” | ”that” | “ though” | ”till” | ”unless” | ”until” | ”when” | ”whenever” | ”where” | ”whereas” | ”wherever” | ”which” | ”while” | ”whilst” | ”who” | ”whoever” | ”whom” | ”whose” | ”as far as” | ”as if” | ((”as” | ”so”) “long as”) | ”as soon as” | ”as though” | (“assuming” [“that”]) | ”considering” | (”given” [“that”]) | (”granted” [“that”]) | ”in case” | ”insofar as” | ”insomuch as” | ”in the event that” | (“providing” | ”provided” [“that”]) | ”seeing as” | (”seeing” [“that”]) | (”so” | ”such” “that”) | (“supposing” [“that”])

<nominal clause> ::= <nominal subordinator> [<noun phrase> | <person>] <verb phrase> \{[<coordinator>] <nominal clause>\}

<nominal subordinator> ::= ”what” | “who” | “whom” | “whose” | “which” | “when” | “where” | “how” | “why” | “because” | “whether”

<relative clause> ::= [<relative subordinator>] (<noun phrase> | <person> | <verb phrase>) [<verb phrase>] \{[<coordinator>] <relative clause>\}

<relative subordinator> ::= ([<preposition>] “which” | “that” | “who” | “whom” | “whose” | “as”) | (“when” | “where” | “why”)

\subsection{Phrase-Level Pattern}

<noun phrase> ::= [<premodifier>] <head> [<postmodifier>] \{[<coordinator> <noun phrase>\}

<head> ::= (<noun> | <pronoun> | <numerals> | <symbol>) \{[<coordinator>] <head>\}

<premodifier> ::= \{<determiner> | <adjective phrase> | <noun>\}

<postmodifier> ::= \{[<noun phrase>] [<relative clause>] [<prepositional phrase>] [<participle phrase>] [<infinitive phrase>] [<adjective phrase>]\}

<verb phrase> ::= [<modal>] [<catenative verb phrase>]
\{[<adverb phrase>] <auxiliary>\} [<adverb phrase>] <lexical verb> [<adverb phrase>] [<noun phrase> | <nominal
clause> | <person>] {<noun phrase> | <person> | <adjective phrase> | <prepositional phrase> | <infinitive phrase> | <participle
phrase> | <relative clause>} \{[<coordinator>] <verb phrase>\}

<participle phrase> ::= \{[<adverb phrase>] <auxiliary>\} [<adverb Phrase>] <participle>  [<adverb phrase>] [<noun phrase> | <nominal clause> | <person>] \{<noun phrase> | <person> | <adjective phrase> | <prepositional phrase> | <infinitive phrase> | <relative clause>\} \{[<coordinator>] <participle phrase>\}

<infinitive phrase> ::= “to” <verb phrase> \{[<coordinator>] <infinitive Phrase>\}

<catenative verb phrase> ::= [<non negative adverb phrase>] (”appear” | “tend” | “seem” | “happen” | “get” | “turn out”) [<non negative adverb phrase>] ”to”

<adjective phrase> ::= [<adverb phrase>] (<adjective> | <past participle> | <present participle>) [<adverb phrase>] [<prepositional phrase> | <infinitive phrase> | <nominal clause>] \{[<coordinator>] <adjective phrase>\}

<adverb phrase> ::= <adverb> \{[<coordinator>] <adverb phrase>\}

<non negative adverb phrase> ::= <non negative adverb> \{[<coordinator>] <non negative adverb phrase>\}

<negative adverb phrase> ::= [<non negative adverb>] <negative adverb>

<prepositional phrase> ::= [<adverb phrase>] <preposition> \{<preposition>\} (<noun phrase> | <person> | <nominal clause> | <verb phrase>) \{[<coordinator>] <prepositional phrase>\}

\subsection{Word-Level Pattern}

The following are closed word class:

<pronoun> ::= “it” | “they” | “one” | “this” | “these” | “that” | “those” | “mine” | “ours” | “yours” | “hers” | “its” | “theirs” | “one’s”

<person> ::= <personal pronoun> | <reflexive pronoun> | <reciprocal pronouns> | <indefinite pronoun>

<personal pronoun> ::= “I” | “we” | “you” | “he” | “she” | “it” | “they” | “one” | “me” | “us” | “him” | “her” | “them”

<reflexive pronoun> ::= “myself” | “ourself” | “yourself” | “yourselves” | “himself” | “herself” | “itself” | “themselves” | “oneself”

<reciprocal pronouns> ::= “each other” | “one another”

<indefinite pronoun> ::= “someone” | “anyone” | “no one” | “everyone” | “somebody” | “anybody” | “nobody” | “everybody” | “something” | “anything” | “nothing” | “everything”

<compound verb> ::= (“come” “into” | “off” | “out” | “up”) | (“get” “at” | “away” | “on”) | (“give” “in” | “off” | “up”) | (“go” “into” | “off” | “up”) | (“hold” “against” | “on” | “to”) | (“keep” “on” | “up” | “to”) | (“knock” “about” | “down” | “over”) | (“let” “off” | “out” | “up”) | (“look” “after” | “forward” | “into” | “over”) | (“make” “for” | “out” | “up”) | (“pick” “for” | “out” | “up”) | (“pick” “on” | “out” | “up”) | (“pull” “over” | “through” | “up”) | (“put” “across” | “forward” | “out”) | (“run” “into” | “over” | “up”) | (“set” “off” | “out” | “up”) | (“take” “back” | “off” | “to”) | (“turn“ “over” | “round” | “up”) | (“work” “on” | “out” | “up”)

<auxiliary> ::= “be” | “have” | “do”

<modal> ::= “can” | “could” | “shall” | “should” | “will” | “would” | “must” | “might” | “may” | “dare” | “need” | “ought to” | “used to”

<negative adverb> ::= "not” | “seldom” | “scarcely” | “never” | “hardly” | “rarely” | “few” | “little” | “nowhere"

<preposition> ::= “aboard” | ”about” | ”above” | “across” | ”after” | ”against” | ”along” | ”amid” | ”among” | ”around” | ”as” | ”at” | ”before” | ”behind” | ”below” | ”below” | ”beneath” | ”beside” | ”besides” | ”between” | ”beyond” | ”but” | ”by” | ”despite” | ”down” | ”during” | ”except” | ”for” | ”from” | ”in” | ”inside” | ”into” | ”like” | ”near” | ”of” | ”off” | ”on” | ”onto” | ”opposite” | ”outside” | ”over” | ”past” | ”round” | ”since” | ”than” | ”through” | ”to” | ”toward” | ”towards” | ”under” | ”underneath” | ”unlike” | ”until” | ”up” | ”upon” | ”via” | ”with” | ”within” | ”without” | “ahead of” | “apart from” | “as for” | “as of” | “as well as” | “because of” | “but for” | “by means of” | “by virtue of” | “due to” | “except for” | “for lack of” | “in addition to” | “in aid of” | “in exchange for” | “in favour of” | “in front of” | “in line with” | “in place of” | “inside of” | “in spite of” | “instead of” | “near to” | “on account of” | “on top of” | “out of” | “outside of” | “owing to” | “prior to” | “subsequent to” | “such as” | “e.g.” | “thanks to” | “up to”

<determiner> ::= ”a few” | ”a little” | ”all” | ”another” | ”any” | ”both” | ”each” | ”either” | ”few” | ”many” | ”much” | ”neither” | ”none” | ”one” | ”several” | ”some” | ”this” | ”that” | ”these” | ”those” | ”my” | ”our” | ”your” | ”his” | ”her” | ”its” | ”their” | ”one’s” | “each other’s” | “one another’s”

<existential there> ::= “there” | ”here”

<coordinator> ::= "and" | "or" | "not only" | "but also" | "both" | "either" | "yet" | "but" | “as well as”

The following are open word class:

<noun>::=  the set of words with the same part-of-speech “noun”.

<numerals> ::= the set of words with the same part-of-speech “cardinal number”.

<symbol> ::= the set of words with the same part-of-speech “symbol”.

<lexical verb> ::= <simple verb> | <compound verb>.

<simple verb> ::= the set of words with the same part-of-speech “verb”, except auxiliary verbs and modal verbs.

<participle> ::= <past participle> | <present participle>.

<past participle> ::= the set of words with the same part-of-speech “past participle”.

<present participle> ::= the set of words with the same part-of-speech “present participle”.

<adjective> ::= the set of words with the same part-of-speech “adjective”.

<adverb> ::= the set of words with the same part-of-speech “adverb”.

<non negative adverb> ::= the set of words with the same part-of-speech “adverb” except the words within <negative adverb>.

\section{SLN-Based Similarity Calculation}
\label{sec:appendix}
The similarity between two phrases is calculated by Algorithm 2.

\begin{algorithm*} \small
    \renewcommand{\thealgocf}{2}
    \caption{Calculate the Similarity of Two Phrases}\label{algorithm}
    \LinesNumbered 
    \KwIn{$phrase_1 = (w_1, w_2, ..., w_i, ..., w_{|phrase_1|})$, $phrase_2 = (w_1^{'}, w_2^{'}, ..., w_j^{'}, ..., w_{|phrase_2|}^{'})$, // each $w_i$ is a word}
    \KwOut{A real value within the range $[0, 1]$}
    \SetKwInOut{KwData}{Function}
    \KwData{
    (1) Usage ($phrase$) // input a phrase and then return a set of phrases represented by $synonym = \{phrase_1,phrase_2,…,phrase_i,…,phrase_{|synonym|}\}$, where each $phrase_i$ denotes a synonymous phrase of the input phrase. (2) ConstructSLN ($phrase$) // input a phrase and then return the corresponding SLN represented by $SLN = (V,L)$, where $V = {v_1,v_2,…,v_i,…,v_{|V|}}$, $L =  {l_1,l_2,…,l_i,…,l_{|L|}}$. Each $v_i$ denotes a noun phrase, and each $l_i=(v_{i1},v_{i2},t_i)$ denotes a $t_i$ type semantic link between $v_{i1}$ and $v_{i2}$. (3) SplitNounPhrase ($phrase$) // according to the pattern of noun phrase, input a noun phrase and then return two items: 1) its premodifier represented by $pre = ((w_1,p_1),(w_2,p_2),…,(w_i,p_i),…,(w_n,p_n))$, where each $w_i$ denotes a word in premodifier and $p_i$ denotes the corresponding part-of-speech, and 2) its head noun represented by $h$.\\
    }
    
    \LinesNumbered 

    \BlankLine
    \SetKwFunction{MyFuns} {ExpandSLN}
    \SetKwProg{Fn}{Function}{:}{}
    \Fn{\MyFuns{$phrase$}} {
        $SLN^{'}$ $\leftarrow$ ConstructSLN ($phrase$)\\
        $V^{'}$, $E^{'}$ $\leftarrow$ $SLN^{'}[0]$, $SLN^{'}[1]$ \\
        $V$, $E$ $\leftarrow$ $\emptyset$, $\emptyset$ // initialize the final SLN to be expanded\\
        // transfer the semantic links between two noun phrases of the original SLN into the links between two corresponding head nouns of the final SLN\\
        \For{$(v_1^{'}, v_2^{'}, t_i^{'})$ $\in$ $E^{'}$}{
            $pre_1$, $h_1$ $\leftarrow$ SplitNounPhrase ($v_1^{'}$) \\
            $pre_2$, $h_2$ $\leftarrow$ SplitNounPhrase ($v_2^{'}$) \\
            $E$ $\leftarrow$ $E$ $\cup$ $(h_1, h_2, t_i^{'})$ // add a link between two head nouns with original type into the final SLN\\
        }
        
        // find the modifying relations within noun phrase, and add them as “mod” link into the final SLN\\
        
        \For{$v^{'}$ $\in$ $V^{'}$}{
            $pre$, $h$ $\leftarrow$ SplitNounPhrase ($v^{'}$)\\
            $V$ $\leftarrow$ $V$ $\cup$ $h$ // add the head noun of a noun phrase in original SLN as a new node into the final SLN\\
            \For{$i$ in range (Len ($pre$))}{
                $pre$, $h$ $\leftarrow$ SplitNounPhrase ($v^{'}$)\\
                $V$ $\leftarrow$ $V$ $\cup$ $h$ // add the head noun of a noun phrase in original SLN as a new node into the final SLN\\
                \For{$i$ in range ($i+1$, Len ($pre$))}{
                    $w_j$, $p_j$ $\leftarrow$ $pre[j][0]$, $pre[j][1]$\\
                    \If{$p_j$ is $Noun$}{
                        $L$ $\leftarrow$ $L$ $\cup$ $(w_i, w_j, "mod")$\\
                    }
                    $L$ $\leftarrow$ $L$ $\cup$ $(w_i, h, "mod")$ // each word in premodifier modifies the head noun\\
                }
            }
        }
        \KwRet {$(V, L)$}
    }

    \BlankLine
    \SetKwFunction{MyFuns} {Similarity}
    \SetKwProg{Fn}{Function}{:}{}
    \Fn{\MyFuns{$SLN_1$, $SLN_2$}} {
        $V_1$, $L_1$, $V_2$, $L_2$ $\leftarrow$ $SLN_1[0]$, $SLN_1[1]$, $SLN_2[0]$, $SLN_2[1]$ \\
        $score$ $\leftarrow$ $0$ // the number of common edges\\
        \For{$(v_1, V_2, t)$ $\in$ $L_1$}{
            $ismatch$ $\leftarrow$ $False$ // whether the current edge in $SLN_1$ exists in $SLN_2$\\
            \For{$(v_1^{'}, V_2^{'}, t^{'})$ $\in$ $L_2$}{
                \If{$(v_1, V_2, t)$ $=$ $(v_1^{'}, V_2^{'}, t^{'})$}{
                    $ismatch$ $\leftarrow$ $1$ \\
                }
            }
            $score$ $\leftarrow$ $score$ + $ismatch$\\
        }
        \KwRet {$result$}
    }

    \BlankLine
    \SetKwFunction{MyFuns} {MatchTwoPhrase}
    \SetKwProg{Fn}{Function}{:}{}
    \Fn{\MyFuns{$phrase_1$, $phrase_2$}} {
        $synonym$ $\leftarrow$ Usage ($phrase_2$) \\
        $maxsimilarity$ $\leftarrow$ $0$ // the maximal similarity between $phrase_1$ and synonymous phrases of $phrase_2$\\
        $SLN_1$ $\leftarrow$ ExpandSLN ($phrase_1$)\\
        \For{$phrase_2^{'}$ $\in$ $synonym$}{
            $SLN_2$ $\leftarrow$ ExpandSLN ($phrase_2^{'}$) \\
            $result$ $\leftarrow$ Similarity ($SLN_1$, $SLN_2$) // this call returns the similarity between $phrase_1$ and one of synonymous phrases of $phrase_2$\\
            \If{$maxsimilarity$ $\leq$ $result$}{
                $maxsimilarity$ $\leftarrow$ $result$
            }
        }
        \KwRet {$maxsimilarity$}
    }
\end{algorithm*}

\section{Previous Patterns}
\label{sec:appendix}
The following patterns were proposed for identifying subclass relation in 1992 \cite{10.3115/992133.992154}:

(1)	(<noun phrase> (((“and” | “or”) “other”) | “especially” | “such as”) <noun phrase>) | (“such” <noun phrase> “as” <noun phrase>). 

(2)	<noun phrase> “including” <noun phrase>.

The above pattern (1) cannot represent definition because (“and” | “or” “other”) | “especially” | “such as” represent examples, e.g., “Besides popular concepts such as "cities" and "musicians" …”, “temples, treasuries, and other important civic buildings.” and “… most European countries, especially France, England, and Spain.“ all represent examples. 

The following definition patterns were proposed in 2000 \cite{10.1145/354756.354817}:

(1)	(<qn> “(” <noun phrase> “)”) | (<noun phrase> “(” <qn> “)”), where qn denotes a specified query noun. 

(2)	<qn> (“is” | “was” | “are” | “were”) (“a” | “an” | “the”) <noun phrase>. 

(3)	<qn> “which” (“is” | “was” | “are” | “were”) <noun phrase>.

The above pattern (1) cannot represent definition because the content in parentheses usually consists of specific instances, e.g., “Understanding short text (e.g., web search, tweets, anchor texts) is important to many applications.”. The above pattern (2) is a special case of our pattern <noun phrase> “be” <noun phrase> after lemmatize “is” and others into “be”, where “a”, “an” and “the” belongs to the premodifier of noun phrase.

The following definition patterns were proposed in \cite{Liu2003MiningTC}:

(1)	(“is” | “are”) [<adverb>] (“called” | “known as” | “defined as”) <concept>, where concept denotes a noun phrase. 

(2)	<concept> ((“refer” | “refers”) “to”) | “satisfy” | “satisfies”.

(3)	<concept> (“is” | “are”) [<adverb>] (“being used to” | “used to” | “referred to” | “employed to” | “defined as” | “formalized as” | “described as” | “concerned with” | “called”).  

(4)	<concept> (“-” | “:”). 

The identifier in the above pattern (3) (“being used to” | “used to” | “employed to”) cannot represent definition because the sentences extracted by them describe that certain measure is taken to achieve certain purpose (e.g., in sentence “[In our method, noun network in WordNet is used] to {extract semantic relatedness for word pairs}.”, the measure “use noun network in WordNet” is beneficial for achieving the purpose “extract semantic relatedness for word pairs”).

The above pattern (4) is insufficient to represent definition because this pattern is relevant to the importance of the concept. The following is an example that non-definition is extracted by using this pattern, “Unfortunately, Yarowsky's method is not well understood from [a theoretical viewpoint] : {we would like to formalize the notion of redundancy in unlabeled data, and set up the learning task as optimization of some appropriate objective function}.”.

The following definition patterns were proposed in 2004 \cite{articleAnswering}:

(1)	<noun phrase> “become” <noun phrase>.

(2)	<noun phrase> [“which” | “that”] <verb phrase>. 

The above pattern (2) cannot represent definition because appositive and relative clause just supplement or constrain a noun phrase in English grammar (e.g., “Another line of work attempts to handle [user queries] which {are ambiguous by asking back clarification questions}.”).

The following definition patterns were proposed in 2013 \cite{Jin2013MiningST}:

(1)	(“define” | “defines”) <term> “as” <definition>, where term denotes a noun phrase to be defined.

(2)	(“definition of” <term> <definition>) | (<term> (“a measure of” | “designate” | “designates”) <definition>).

(3)	<term> (“comprise” | “comprises” | (“consist” |“consists” “of”)) <definition>.

(4)	<term> (“denote” | “denotes”) <definition>.

The identifiers of above patterns can be summarized as: 

<previous identifier> ::= “including” | ((“is” | “was” | “are” | “were”) (“a” | “an” | “the”)) | (“which” (“is” | “was” | “are” | “were”)) | ((“is” | “are”) [<adverb>] (“called” | “known as” | “defined as”)) | ((“refer” | “refers”) “to”) | (“is” | “are”) [<adverb>] (“referred to” | “defined as” | “formalized as” | “described as” | “called”) | “become” | (“comprise” | “comprises” | (“consist” |“consists” “of”)) | (“denote” | “denotes”)

These patterns are used as a baseline to compare in our self-built dataset and public datasets.

Comparing with the previous patterns, this research proposes the following new definition patterns:

(1)	 Pattern with synonymy identifier: <noun phrase> [“which” | “that”] (“be” (“to” | “that”)) | (("represent" | ”indicate” | ”mean”) [“to” | “that”]) | “namely” | ([“be”] ("equal" | “equivalent” | ”correspond” | “amount” | ”refer”) “to”) | ([“be”] (“denote” | "represent" | ”indicate” | ”mean” | ”redefine” | ”express” | “give”) (“as” | “by”)) | ([“be”] (“introduce” | ”propose” | ”report” | ”develop” | ”design” | ”articulate” | “view” | ”employ” | “use” | “reuse” | “utilize” | “think” | “see” | ”regard” | “consider” | “count” | “select”) “as”) (<noun phrase> | <nominal clause> | <verb phrase> | <main clause>)

(2)	Pattern with part-whole identifier: <noun phrase> “have” <noun phrase>,
<noun phrase> [“which” | “that”] “have” | ("contain" | "involve" | "include" | "encompass") | ([“be”] (“make up" | "compose" | "constitute") (“of” | ”from”)) | ([“be”] (“decompose" | "split" | "break" | "divide" | "separate" | "categorize" | "classify") (“in” | ”to” | ”into” | ”between” | “among”)) (<noun phrase> | <participle phrase> | <nominal clause>), and
(<noun phrase> | <participle phrase> | <nominal clause>) [“which” | “that”] (“make up" | "compose" | "constitute") | ([“be”] (“consist" | "contain" | "involve" | "include" | "comprise" | "encompass") (“in” | ”to” | ”into”)) |  ([“be”] (“add" | "supplement" | ”group” | ”combine”) (“in” | ”to” | ”into” | ”between” | “among”)) <noun phrase> | <nominal clause> | <verb phrase> | <main clause>)

In addition, the preprocess of input enables our pattern to cover more cases.

\section{Patterns with different precisions and recalls}
\label{sec:appendix}
Table 6 shows the low-level patterns of definition with different precisions and recalls based on experiments on self-built dataset.

\begin{table*}
\centering
\begin{tabular}{p{12.5cm}ll}
\hline
Pattern & Precision & Recall\\
\hline
<noun phrase> [“which” | “that”] "be" (“a” | “an”) [(“kind” | <synonym of kind>) “of”] <noun phrase> & 91.5 & 83.1\\
<noun phrase> [“which” | “that”] “have” <noun phrase> & 94.7 & 12.9\\
<noun phrase> [“which” | “that”] ("consist of” | <synonym of consist of>) (<noun phrase> | <participle phrase> | <nominal clause>) & 98.7 & 9.8\\
<noun phrase> [“which” | “that”] ([“be”] (“make up" | <synonym of make up>) (“of” | ”from”)) (<noun phrase> | <participle phrase> | <nominal clause>) & 99.0 & 1.0\\
<noun phrase> [“which” | “that”] ([“be”] (“decompose" | <synonym of decompose>) (“in” | ”to” | ”into” | ”between” | “among”)) (<noun phrase> | <participle phrase> | <nominal clause>) & 98.9 & 1.8\\
(<noun phrase> | <participle phrase> | <nominal clause>) [“which” | “that”] (“make up" | <synonym of make up>) <noun phrase> & 97.4 & 0.8\\
<noun phrase> [“which” | “that”] "be" (“a” | “an”) [(“kind” | <synonym of kind>) “of”] <noun phrase> & 91.5 & 83.1\\
(<noun phrase> | <participle phrase> | <nominal clause>) [“which” | “that”] ([“be”] (“add" | <synonym of add>) (“in” | ”to” | ”into” | ”between” | “among”)) <noun phrase> & 96.0 & 1.9\\
<noun phrase> [“which” | “that”] “be” <noun phrase> & 88.5 & 32.8\\
<noun phrase> [“which” | “that”] (“be” (“to” | “that”)) | (("denote" | <synonym of denote>) [“to” | “that”]) | “namely” | “become” (<noun phrase> | <nominal clause> | <verb phrase> | <main clause>) & 94.6 & 8.4\\
<noun phrase> [“which” | “that”] ([“be”] (“equivalent” | <synonym of equivalent>) “to”) (<noun phrase> | <nominal clause> | <verb phrase> | <main clause>) & 96.4 & 3.5\\
<noun phrase> [“which” | “that”] ([“be”] “call” | “name” [“as”]) | ([“be”] ("denote" | <synonym of denote> | “define” | <synonym of define>) (“by” | “as”)) | ([“be”] (”introduce” | <synonym of introduce> |  “view” | <synonym of view>) “as”) (<noun phrase> | <nominal clause> | <verb phrase> | <main clause>) & 90.4 & 17.0\\
(“call” | “name” | “define” | <synonym of define> | “introduce” | <synonym of introduce> | “view” | <synonym of view>) <noun phrase> (<noun phrase> | <relative clause>) & 90.4 & 10.5\\
\hline
\end{tabular}
\caption{\label{citation-guide}
Patterns of definition with different precisions and recalls.
}
\end{table*}

Table 7 shows the low-level patterns of method with different precisions and recalls based on experiments on self-built dataset.

\begin{table*}
\centering
\begin{tabular}{p{12.5cm}ll}
\hline
Pattern & Precision & Recall\\
\hline
<noun phrase> [“which” | “that”] "be" (“a” | “an”) [(“kind” | <synonym of kind>) “of”] <noun phrase> & 91.5 & 83.1\\
<noun phrase> [“which” | “that”] “have” <noun phrase> & 94.7 & 12.9\\
<noun phrase> [“which” | “that”] ("consist of” | <synonym of consist of>) (<noun phrase> | <participle phrase> | <nominal clause>) & 98.7 & 9.8\\
<noun phrase> [“which” | “that”] ([“be”] (“make up" | <synonym of make up>) (“of” | ”from”)) (<noun phrase> | <participle phrase> | <nominal clause>) & 99.0 & 1.0\\
<noun phrase> [“which” | “that”] ([“be”] (“decompose" | <synonym of decompose>) (“in” | ”to” | ”into” | ”between” | “among”)) (<noun phrase> | <participle phrase> | <nominal clause>) & 98.9 & 1.8\\
(<noun phrase> | <participle phrase> | <nominal clause>) [“which” | “that”] (“make up" | <synonym of make up>) <noun phrase> & 97.4 & 0.8\\
<noun phrase> [“which” | “that”] "be" (“a” | “an”) [(“kind” | <synonym of kind>) “of”] <noun phrase> & 91.5 & 83.1\\
(<noun phrase> | <participle phrase> | <main clause>) [“which” | “that”] [“be”] (“rely” | <synonym of rely>) ("in" | "on" | "upon" | "around") (<noun phrase> | <verb phrase>) | ([“be”] (“rely” | <synonym of rely>) ("in" | "on" | "upon" | "around")) (<noun phrase> | <verb phrase>) <main clause>
 & 98.1 & 5.7\\
(<noun phrase> | <participle phrase> | <main clause>) [“which” | “that”] (“require” | <synonym of require>) (“to” | ”for” | ”in” | “at”| “of”) (<noun phrase> | <verb phrase>) & 83.5 & 9.8\\
\hline
\end{tabular}
\caption{\label{citation-guide}
Patterns of method with different precisions and recalls.
}
\end{table*}

\section{Automatic Extraction}
\label{sec:appendix}
Algorithm 1 for extracting definitions and methods consists of the following steps:

(1)	Every sentence is transformed into constituency tree which represents the syntactic structure of the sentence.

(2)	Enumerate every node in the constituency tree to find the identifier defined in the pattern. Choose the longest identifier if two or more identifiers are detected.

(3)	If the nodes adjacent to the identifier are in {<noun phrase>, <verb phrase>, <participle phrase>, <main clause>, <relative clause>, <nominal clause>} defined in the patterns, the sentence is definition or method.

\section{Extraction of Definiendum, Definiens, Measure and Purpose}
\label{sec:appendix}
Within a single sentence, a definition can be represented by a pair of definiendum and corresponding definiens, a method can be represented by a pair of measure and corresponding purpose. The pair annotated manually is represented by $([la1,ra1],[la2,ra2])$, and the pair extracted by patterns is represented by $([le1,re1],[le2,re2])$, where $l$ and $r$ respectively represent the leftmost position and rightmost position of a consecutive sequence of words. An annotated pair and an extracted pair is matched if and only if $la1=le1 \land ra1=re1 \land la2=le2 \land ra2=re2$.

\begin{table}[H] \small
\centering
\begin{tabular}{lllll}
\hline
Type & Annotated & Precision & Recall & F1-Score\\
\hline
Definition & 12682 & 83.1 & 85.0 & 84.0 \\
Method & 18213 & 80.1 & 81.3 & 80.7 \\
\hline
\end{tabular}
\caption{\label{citation-guide}
Experiment result for extracting definiendum, definiens, measure and purpose.
}
\end{table}

\end{document}